\pdfoutput=1
\documentclass[times,referee,twocolumn,final,authoryear]{elsarticle}

\usepackage{ycviu}
\usepackage{framed,multirow}
\usepackage{tikz}
\usepackage{amssymb}
\usepackage{amsmath}
\usepackage{latexsym}
\usepackage{adjustbox}
\usepackage{listings}
\usepackage{subcaption}
\usepackage[title,titletoc,page]{appendix}
\usepackage{url}
\usepackage{xcolor}
\definecolor{newcolor}{rgb}{.8,.349,.1}

\journal{Computer Vision and Image Understanding}

\begin{document}
\begin{frontmatter}

\title{}

\title{PS-DeVCEM: Pathology-sensitive deep learning model for video capsule endoscopy based on weakly labeled data}
\author[1]{Ahmed \snm{Mohammed}\corref{cor1}}
\cortext[cor1]{Corresponding author: 
	Tel.: +47-486-766-16}
\ead{mohammed.kedir@ntnu.no}

\author[1]{Ivar \snm{Farup}}
\author[1]{Marius \snm{Pedersen}}
\author[2]{Sule \snm{Yildirim}}
\author[3]{{\O}istein \snm{ Hovde}}
\address[1]{Department of Computer Science, Norwegian University of Science and Technology, Gj{\o}vik, Norway}
\address[2]{Department of Information Security and Communication Technology, Norwegian University of Science and Technology, Gj{\o}vik, Norway}
\address[3]{Department of gastroenterology, Innlandet Hospital Trust, Gj{\o}vik and Institute of Clinical Medicine, University of Oslo.}

\received{1 May 2013}
\finalform{10 May 2013}
\accepted{13 May 2013}
\availableonline{15 May 2013}
\communicated{S. Sarkar}

\begin{abstract}
We propose a novel pathology-sensitive deep learning model (PS-DeVCEM) for frame-level anomaly detection and multi-label classification of different colon diseases in video capsule endoscopy (VCE) data.
Our proposed model is capable of coping with the key challenge of colon apparent heterogeneity caused by several types of diseases. 
Our model is driven by attention-based deep multiple instance learning and is trained end-to-end on weakly labeled data using video labels instead of detailed frame-by-frame annotation.
This makes it a cost-effective approach for the analysis of large capsule video endoscopy repositories.
Other advantages of our proposed model include its capability to localize gastrointestinal anomalies in the temporal domain within the video frames, and its generality, in the sense that abnormal frame detection is based on automatically derived image features. The spatial and temporal features are obtained through ResNet50 and residual Long short-term memory (residual LSTM) blocks, respectively.
Additionally, the learned temporal attention module provides the importance of each frame to the final label prediction.
Moreover, we developed a self-supervision method to maximize the distance between classes of pathologies.
We demonstrate through qualitative and quantitative experiments that our proposed weakly supervised learning model gives a superior precision and F1-score reaching, 61.6\% and 55.1\%, as compared to three state-of-the-art video analysis methods respectively.
We also show our model’s ability to temporally localize frames with pathologies, without frame annotation information during training.
Furthermore, we collected and annotated the first and largest VCE dataset with only video labels.
The dataset contains 455 short video segments with 28,304 frames and 14 classes of colorectal diseases and artifacts.
Dataset and code supporting this publication will be made available on our home page.
\end{abstract}


\end{frontmatter}

\section{Introduction}
There are several colorectal diseases and abnormalities that interfere with the normal working of the colon. 
Colorectal diseases include colorectal cancer, polyps, ulcerative colitis, diverticulitis etc. 
Screening and detection of colorectal diseases at an early stage could improve disease management and diagnosis.
Colonoscopy is considered a gold standard as a screening procedure for colorectal diseases.
However, colonoscopy has some limitations including invasiveness, discomfort, and embarrassment for the patient and relatively
high cost. 
These inconveniences may limit the utility of colonoscopy, especially in screening strategies where acceptance of the test is of the utmost importance~(\cite{schoofs2006pillcam}).
VCE is an alternative to visualize the colon.
VCE was first introduced in 2006 as a wireless, minimally invasive technique for the imaging of the large bowel that does not require sedation or gas insufflation~(\cite{shi2015role}).
A single VCE procedure produces approximately 50,000 images and takes 45-90 minutes to review. 
Therefore, a machine learning system can be used to complement gastroenterologists for fast and accurate diagnosis~(\cite{li2011computer}).   

The detection and classification of colorectal diseases is a very challenging problem due to apparent colon heterogeneity.
In fact, colon data contains a high degree of apparent heterogeneity due to varying levels of unpredictable responses caused by the nature of diseases.
Moreover, the morphological clues in local neighborhoods of colon images are not consistent across different patients. This makes it difficult to develop an automated disease detection model based on image analysis techniques. 
A number of image processing based methods have addressed colorectal pathology detection problems in literature ~(\cite{mohammed2018net,bernal2017comparative,tajbakhsh2015,ronneberger2015u}).
These methods do not consider long-term temporal dependencies between frames to improve the performance of detection algorithms.  
Moreover, they rely on the assumption that pixel-level or frame-level annotation data is available and is trained in a fully supervised manner.
However, this assumption is very limiting in a clinical setting as it is expert-intensive and time-consuming to obtain a precise annotation of the pathologies per image.
In addition, from a clinical application perspective, a gastroenterologist is required to check for various types of pathologies during a single examination and computer-aided diagnostic techniques are expected to detect as many diseases as possible to circumvent miss-diagnosis.
However, the number of pathologies that earlier methods handle is limited to classes of diseases such as polyps~(\cite{bernal2017comparative}) and angiodysplasia~(\cite{shvets2018angiodysplasia}).
Moreover, the dataset used in training such models lacks class variety to be used in clinical application.

\begin{figure*}[ht]
	\centering
	\includegraphics[width=1\linewidth]{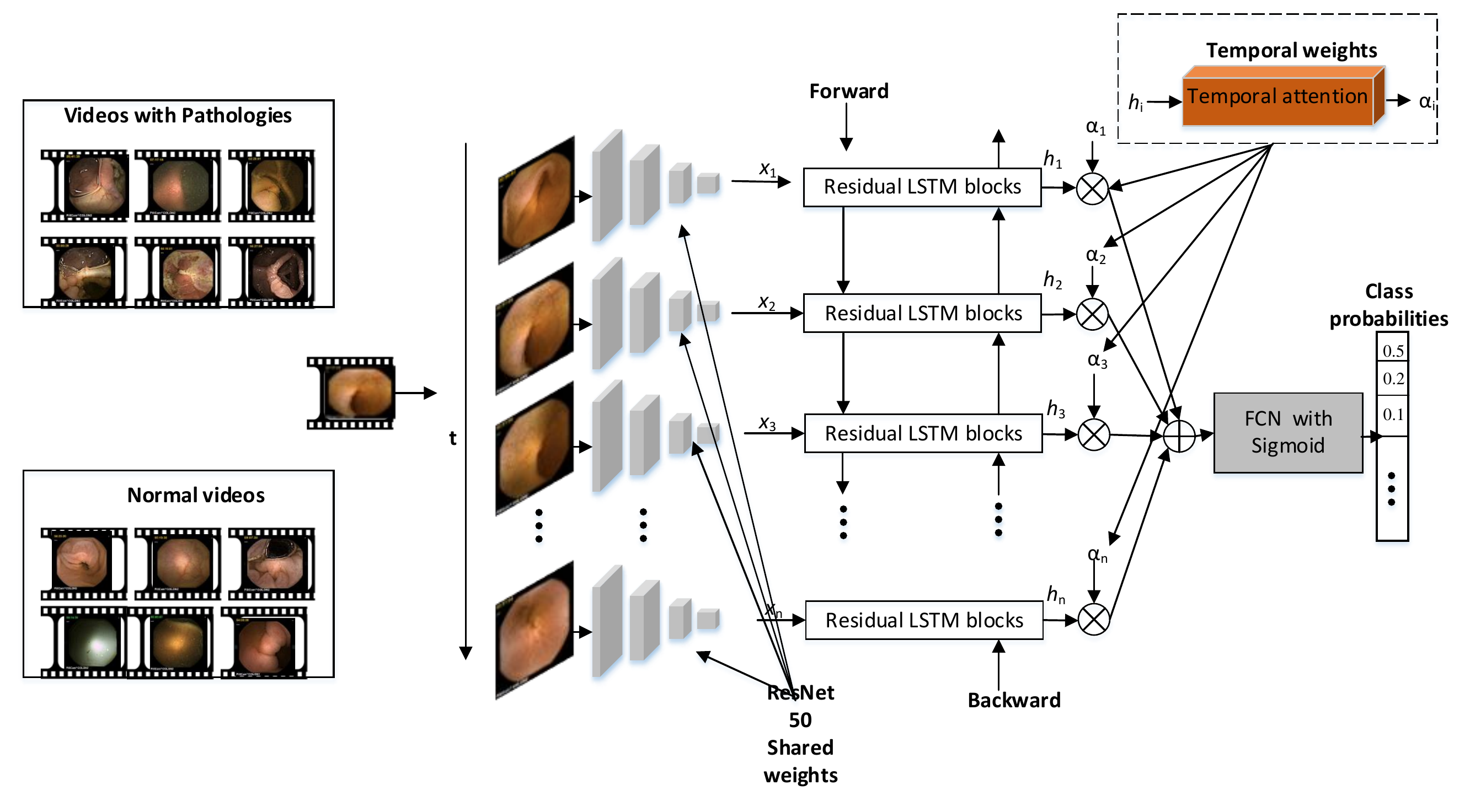}
	\caption{PS-DeVCEM: The frame features are extracted with ResNet50 network~(\cite{he2016deep}) pre-trained on ImageNet~(\cite{deng2009imagenet}). The feature embedding is computed by passing through a residual LSTM block. Finally, the embeddings are aggregated with learned weights. The output of the network is video-level class probabilities for each pathology. For details please refer to Sec. 3.}
	\label{fig:teaser}
\end{figure*}

To address these challenges, we propose PS-DeVCEM, a new weakly supervised learning approach for learning frame-level multi-label classification from a given video label. 
Our model explores robust deep residual features that are invariant to apparent heterogeneity in colon data.
We also explore residual LSTM units to take into account the long term photo-metric and appearance variability. 
Our approach is based on an objective function that minimizes within-video similarities between positive and negative feature frames, while at the same time learning video-level prediction and contribution from each frame.
The proposed method requires only video labels which can easily be obtained from VCE data reader software tags such as RAPID reader~(\cite{RAPIDRe77:online}) as a normal working procedure.
We formulate the aggregation of positive and negative frame labels using the Bernoulli distribution.
The network is trained by optimizing the sum of log-likelihood and cross-entropy for the video prediction.
The final video prediction is computed by taking the learned weighted mean of each frame feature embeddings.
The learned weights are given by a two-layered neural network that corresponds to the attention.
High attention weights indicate key-frames for the detected pathology class and low attention weights show non-informative frames.
We exploit the ordering of the attention weights to minimize the similarities between high and low attention frames by training a two-layered neural network which acts as a self-supervision method. 
The flow diagram of our proposed PS-DeVCEM model is shown in Fig. (\ref{fig:teaser}).

Moreover, we assume the video is permutation-variant and modeled with residual bi-directional LSTM.
Furthermore,  we collect short VCE video segments. The goals are two-fold; to provide a research and development resource for VCE pathology detection, and to provide a way to benchmark and compare different approaches.
The dataset contains a total of 28,304 video frames containing 14 classes of diseases and artifacts (bubble and debris) from 40 patients.

The primary contributions of this paper are:
\begin{itemize}
	\item  A self-supervision method to minimize the similarity between positive and negative frames within a video segment. This has an advantage by forcing the attention network to be discriminative between positive and negative instances.
	\item An end-to-end trainable network that takes video as an input and detects key-frames with video-level prediction.
	The features from each frame in the video are aggregated using learned weights for final video prediction.
	In addition, we assume temporal dependency between neighboring frames, which is modeled using residual bidirectional long short-term memory blocks.
	\item A new VCE dataset suited for weakly supervised learning problems. The dataset contains 455 short video segments extracted using RAPID reader software~(\cite{RAPIDRe77:online}). There are 28,304 frames with a total of 14 classes of diseases from 40 patients.
	\item Detailed comparison of existing weakly supervised learning algorithms~(\cite{ray2005supervised,andrews2003support,zhou2007relation,bunescu2007multiple,ilse2018attention,paul2018w,nguyen2018weakly}) on VCE dataset. We employ a random 50-50 train/test split under the condition that at least each pathology exists in the train/test set.  
	\item Qualitative and quantitative experiments on VCE dataset and comparison with state-of-the-art works show that our proposed method outperforms in-terms of precision and F1-score reaching 61.6\% and 55.1\% respectively.
\end{itemize}

We organize the rest of the paper as follows. Section 2 briefly reviews previous works on video analysis and multiple instance learning (MIL). In Section 3 we present the PS-DeVCEM along with the self-supervising method. In Section 4 we present the dataset and comparison of different MIL methods and benchmarks will be discussed. In addition, we present experiments with a different configuration of the proposed method. Finally, in Section 5 we conclude the paper with future direction and discussion.
\section{Related Work}
In general, there are two approaches in modeling video data context: short and long context modeling.
In these methods, the long and short-range dependencies can be well memorized by sequentially running the network over individual frames.
However, designing an architecture for video analysis is a challenging task as it involves computationally expensive tasks such as temporal information fusion strategy, frame feature representation (as compared to end-to-end training) and spatio-temporal feature fusion. 
The basic building blocks for video analysis with deep learning includes spatial feature extraction unit such as ResNet~(\cite{he2016deep}), VGG~(\cite{simonyan2014very}), etc. and temporal feature extraction unit such as optical flow and LSTM~(\cite{graves2013generating}) units.
LSTM~(\cite{graves2013generating}) is combined with CNN for activity recognition in Long-term recurrent convolutional networks for visual recognition and description~(\cite{donahue2015long}). 
Other alternative approaches extract spatio-temporal features together using 3D convolutions such as C3D~(\cite{tran2015learning}). Spatio-temporal features such as C3D are used in ~(\cite{sultani2018real}) for anomaly detection in natural videos.
Most of the current state of the art methods use two-stream networks such as ActionVLAD~(\cite{girdhar2017actionvlad}) at the expense of high computational complexity. This is usually done by fusing extracted spatial and optical flow features independently.

In many endoscopic pathology detection problems, labels are relatively scarce and expensive to obtain.  One such case is in VCE, where annotating pathologies frame by frame is arduous and time consuming for medical doctors.
Therefore, weakly supervised approaches such as MIL~(\cite{maron1998framework,mollersen2018bag}) or fully unsupervised methods of detection and segmentation are required to address the above issue.
MIL is a type of weakly supervised learning problem where only group-level, also known as bag level annotation, is available.
The instances within the bag are not labeled. For example, the annotation could be a general statement about the category of the pathology in the video without information about the location within the video or frame labels.
In the MIL problem formulation~(\cite{ilse2018attention}, it is assumed that positive bag videos contain at least one instance of a given pathology while a negative bag video depicts none.

MIL algorithms can be divided into two categories, depending on if the data is an independent samples (images) or temporal based (video).\newline 
\textbf{Independent samples (images):} Assumes the data within a positive or negative bag is an independent sample. The simplest approach to MIL is single-instance learning (SIL)~(\cite{ray2005supervised}) which assigns each instance the label of its bag, creating a supervised learning problem, but mislabeling negative instances in positive bags~(\cite{doran2014theoretical}). In~(\cite{andrews2003support}), the standard support vector machine (SVM) formulation~(\cite{suykens1999least}) is modified so that the constraints on instance labels correspond to the fact that at least one instance in each bag is positive~(\cite{doran2014theoretical}). Similarly, in~(\cite{bunescu2007multiple}), SVM formulation is modified assuming there are very few positive instances of the positive bags. Other unsupervised methods to MIL include MissSVM~(\cite{zhou2007relation}). More recently, Ilse et al.~(\cite{ilse2018attention}) proposed a permutation-invariant aggregation operator that corresponds to the attention method. Compared to soft attention method as in~(\cite{xu2015show}), the aggregation operator is different in that the former is calculated as a dot product while the latter is computed using a two-layered neural network. Moreover, the aggregation operator outperforms commonly used MIL pooling operators~(\cite{ilse2018attention}). In comparison to other works in endoscopy,~(\cite{wang2016computer}) addresses endoscopic images with MIL formulation. Wang et al.~(\cite{wang2016computer}) proposed using endoscopic images with weak labels mined from the diagnostic text. If the diagnostic text does not
match any of predefined of key words such as  Gastric Cancer, Esophageal Cancer, and Esophagitis, the corresponding label of the endoscopic image folder is annotated as negative; otherwise, the
label of the endoscopic image folder is annotated as positive. Each frame is considered as independent. 
\newline
\textbf{Temporal based (video) MIL:} Kotzias et al.~(\cite{kotzias2015group}), proposed using group-level labels to learn instance-level classification models. The group-level prediction is given by taking the average of the instances. An objective function is introduced to encourage smoothness of inferred instance-level labels based on instance-level similarity, while at the same time respecting group-level label constraints. Unlike Kotzias et al. ~(\cite{kotzias2015group}), MI-Net~(\cite{wang2018revisiting}) does not rely on inferring instance-level probabilities. Both of the above approaches are based on neural networks, but Mi-Net is an embedded space aggregation method that uses the MIL pooling layer to focus on learning bag representation. MIL pooling~(\cite{pinheiro2015image}) layer is used to aggregate instance features into one bag representation. Finally, a fully connected (FCN) layer with sigmoid is used to predict the bag labels. In~\cite{paul2018w}, spatial and temporal features are extracted using a two-stream network ( RGB streams and optical flow) and co-activity similarity loss is proposed to maximize the distance between multiple activities. In~(\cite{nguyen2018weakly}) they consider the problem of untrimmed videos by extracting segment features and sparsity loss on the attention weights for aggregating the segment features.

In general, image-based MIL approaches do not provide temporal localization for the detected pathology and are not suitable for VCE video analysis. In VCE or medical imaging applications in general, experts are interested to know frame-level pathology prediction of the detection algorithm. Compared to~(\cite{wang2016computer}) endoscopic images are very different from VCE images. VCE images do not have as good image quality as the traditional endoscopy because of high compression and low image resolution due to volume and power limitation. Bad imaging conditions such as low illumination, uncontrolled capsule motion, and peristalsis, will further reduce the qualities of VCE images.
Among the temporal and independent sample  MIL formulation, none of the above methods exploit the positive and negative segments within a single bag to maximize the distance between the classes. In the proposed method, we show that by using within bag similarity as self-supervision, we can boost the performance of frame localization and VCE video classification.

\section{Pathology-sensitive deep learning model}
Our aim is to design a weakly supervised model for the purpose of multi-label pathology detection.
The model consists of fundamental CNN pipelines, attention, residual LSTM, and self-supervision submodule, as the Fig. (\ref{fig:teaser}) shows. 
The advantage of our attention mechanisms is that it can identify suspected frames and provides a robust video feature representation, while likewise suppressing the irrelevant and non-informative video frames from other classes.
Hence, it is very applicable to weakly supervise learning.
The residual LSTM submodule is able to focus on temporal features among a long sequence of video frames, and while filter out irrelevant features for representation.
Besides, we propose a novel self-supervision mechanism which is used for robust frame localization, because of the apparent colon heterogeneity of the weakly labeled video is quite difficult to distinguish.

We begin by formally defining MIL, and establishing the notation that will be used in the rest of this paper. Let $V=\{f_1, f_2, f_3,..., f_N\}$ be a video containing frames $f_1, f_2, f_3, ..., f_N$ and $N$ is the number of frames in the video. We assume individual labels are available for each video $V$ and is given by $G$ with unknown frame label $\boldsymbol{y}=\{y_1,y_2,y_3,\ldots, y_N\}$. Earlier works in MIL assume binary classification where $y_n \in \{0,1\}$~(\cite{ilse2018attention}),(\cite{wang2018revisiting}). But here we assume a general multi-label classification problem where $y_n$ can assume a set of all possible classes $k$, $P=\{p_1,p_2,p_3,\ldots, p_k\}$ in a multi-label learning problem. $P_k$ is defined as $k^{th}$ abnormality in the dataset and a given video could be labeled to contain multiple abnormalities such as $P=\{``polyp'', ``bleeding'', ``errosion''\}$.
 Hence the ground truth dataset has the form $\mathcal{D}=\{(\boldsymbol{V_1}, Y_1),\ldots, (\boldsymbol{V_n}, Y_n)\}$ where $\boldsymbol{V_i} \in \mathcal{V}$ and ${Y_i} \subseteq {P}$. 
Using the above notation, the MIL constraints could be represented as:
\begin{equation}
\label{eq:1}
Y= 
\begin{cases}
\boldsymbol{p}   & \text{if } \exists  n \quad s.t.\quad  y_n= \boldsymbol{p} ,  \boldsymbol{p} \subseteq P, n \in N \\
0,              & \text{otherwise}
\end{cases}
\end{equation}
where $Y$ is the predicted video label. Alternative MIL constraint formulation can be given as the maximum class probability over the frames as:
\begin{equation}
\label{eq:2}
Y= \max\limits_{ n}\{y_n\} \mid \quad y_n= \boldsymbol{p} ,  \boldsymbol{p} \subseteq P, n \in N
\end{equation}

It is important to note that the frame-level labels, $y_n$ are not available during the training phase and only the video label $G$  is provided. Therefore, our goal is to infer video label $Y$ and frame label $y_n$ by propagating information from video-level to frame-level with a neural network. The motivation for using neural networks is that it is easier to train in an end-to-end fashion. Moreover, previous works~(\cite{ilse2018attention,wang2018revisiting,kotzias2015group,wu2015deep}) have shown that neural network-based MIL approach gives promising results compared to classical approaches~(\cite{andrews2003support,ray2005supervised}).
\subsection{Residual LSTM}
\begin{figure*}[h]
	\centering
	\includegraphics[width=0.6\linewidth]{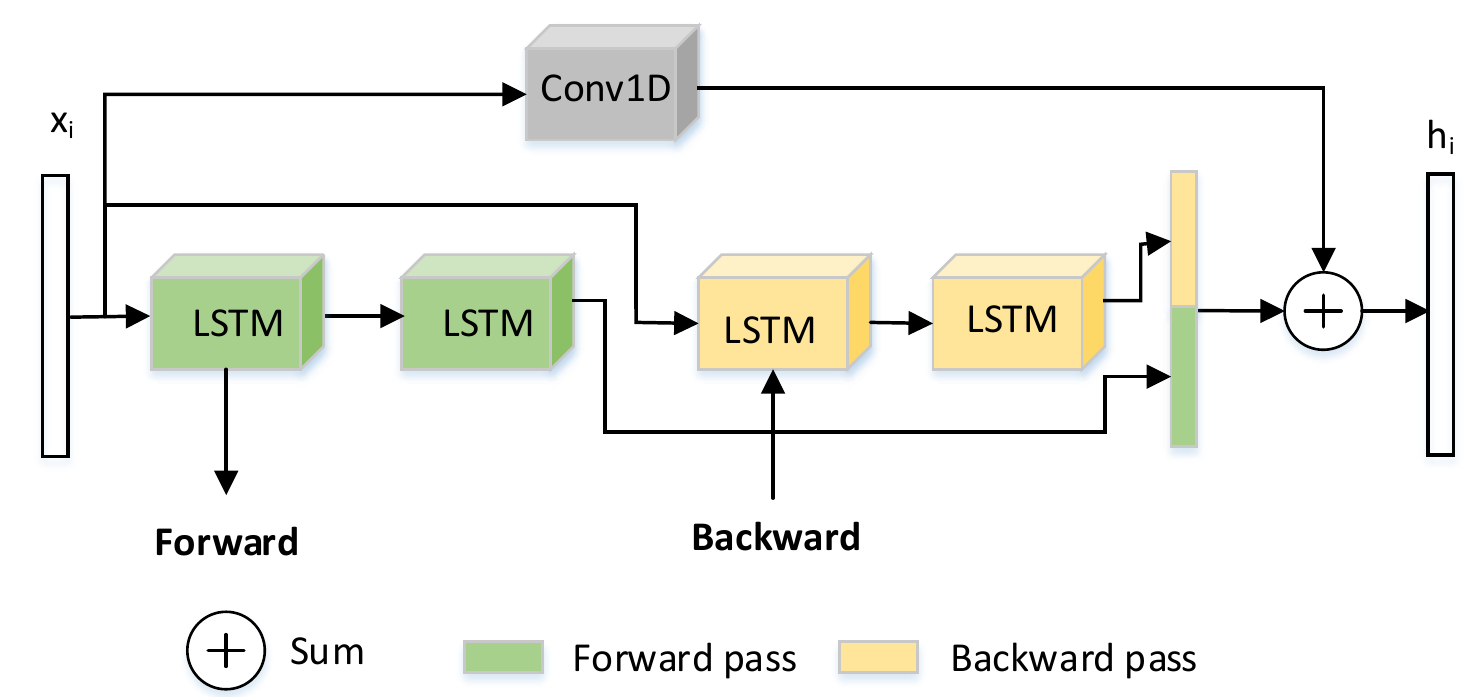}
	\caption{Residual LSTM block. The input to the residual LSTM blocks are spatial features extracted using cascade of convolutions. The forward and backward LSTM temporal informations are concatenated which represents the temporal feature. The temporal features are summed together with spatial input features $x_i$, after passing through a $1x1$ convolution to have the same feature dimension with temporal features.}
	\label{fig:fig2}
\end{figure*}

There are three different approaches to come up with a video-level feature representation.
These are instance aggregation approach~(\cite{andrews2003support}), group aggregation approach~(\cite{cheplygina2015multiple}) and embedded space aggregation approach~(\cite{ilse2018attention}).
The approaches differ in whether they estimate frame-level probabilities or aggregate the embeddings.
Instance aggregation approach works by combining instance-level predictions while group-level aggregation approaches use group similarity for clustering positive and negative samples.
Embedded space aggregation approaches merge instant features and learn group-level classifier~(\cite{wang2018revisiting}).
In VCE or medical imaging applications in general, experts are interested to know frame-level pathology prediction of detection algorithm more than video label-level predictions.
Hence, instance aggregation approaches are suitable for a medical application.  This is because frame-level predictions are paramount as it gives interpretation and explanation for the video prediction.
Our approach is based on an aggregation of embeddings with learned aggregation weight, i.e. attention, which gives frame-level inference to the final video prediction. The framework (illustrated in Fig. \ref{fig:teaser}) consists of $N$ fully convolution encoder networks which extract features $x_i=\Phi_\theta(f_i)$, for each frame. The encoder network $\Phi_\theta$ is ResNet50~(\cite{he2016deep}) that is pre-trained on ImageNet~(\cite{deng2009imagenet}). However, it is possible to use other networks such as VGG~(\cite{simonyan2014very}), DenseNet~(\cite{huang2017densely}) or similar architectures. Temporal dependency between each frame is modeled using residual LSTM blocks as shown in Fig. \ref{fig:fig2}. The residual LSTM blocks consist of bi-directional LSTMs composed of two LSTM units that leverage the residual connection~(\cite{graves2005framewise,hochreiter1997long}). The main idea for using residual connection is to make training easier and avoid performance degradation in deeper networks~(\cite{he2016deep}). The biggest advantage of bi-directional LSTM networks lies in their capability of preserving information over time by the recurrent method. 

\subsection{Temporal attention}
\begin{figure*}[h]
	\centering
	\includegraphics[width=0.6\linewidth]{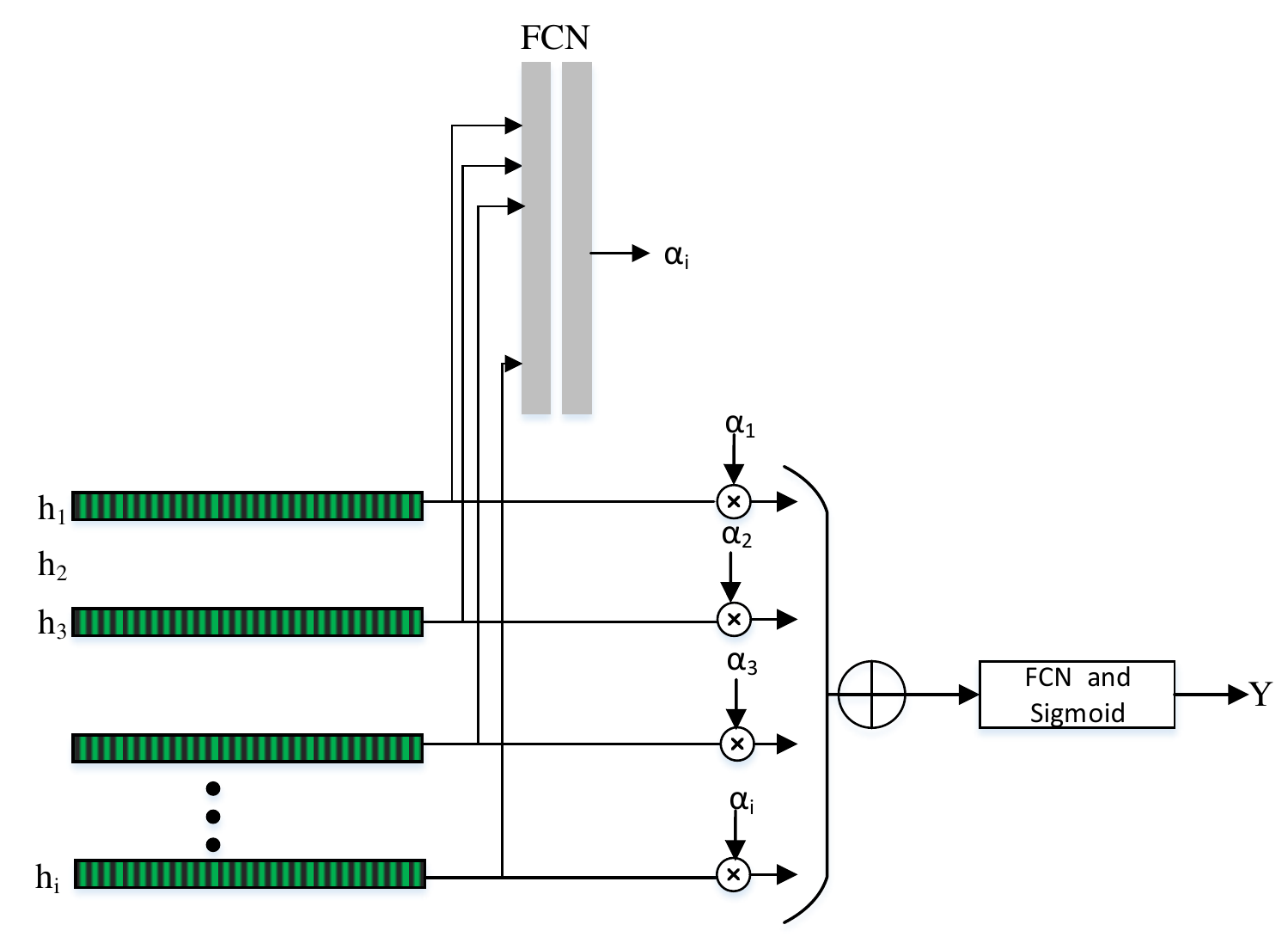}
	\caption{Temporal attention: Given the spatial and temporal feature as input from our residual LSTM block, the temporal attention block computes the relevance of each frame for the final video feature representation. Once the attention values $\alpha_i$ for each frame is computed, the summed weighted features are passed through a fully connected network for video level prediction.}
	\label{fig:fig3}
\end{figure*}

Attention has been shown to improve performance of recurrent neural networks in language translation~(\cite{vaswani2017attention}) and activity recognition~(\cite{sharma2015action}) tasks. Attention is mainly used for easier modeling of long-term dependencies. However, the application of attention to MIL is mainly to model MIL pooling and has been limited~(\cite{ilse2018attention}. Inspired by~(\cite{ilse2018attention}), the temporal attention is parameterized using a two layered neural network. The attention block is shown in Fig. (\ref{fig:fig3}). However, as shown in Fig. (\ref{fig:teaser}) the attention block is trained on residual temporal features rather than frame feature $x_i$ as in~(\cite{ilse2018attention,raffel2015feed}).

The MIL pooling operator aggregates activations of feature maps $h_i$ of the residual block. Let $H=\{h_1,h_2,h_3,...,h_N\}$ be frame representation of the residual block. Hence, the MIL pooling layer is given by~(\cite{ilse2018attention}):
\begin{equation}
\label{eq:3}
Z= \sum_{n=1}^{N} \alpha_nh_n
\end{equation}
where
\begin{equation}
\label{eq:4}
\alpha_n= \frac{\exp\{\boldsymbol{w}^T \text{tanh}(\boldsymbol{Vh}_n^T)\}}{\sum_{i=1}^{N} \exp\{\boldsymbol{w}^T \text{tanh}(\boldsymbol{Vh}_i^T)\} }
\end{equation}
where $\boldsymbol{w} \in \mathbb{R}^{L \times 1}$ and $\boldsymbol{V} \in \mathbb{R}^{L \times M}$ are parameters of two-layered neural network. Such formulation allows the gradient of cost function to be back-propagated efficiently as 'tanh' supports both positive and negative values.  In this formulation, attention can
be seen as producing relative informativeness of the input feature by computing an adaptive weighted average of the residual features. 
\subsection{Self-supervision}
\label{sec:selfsuper}
\begin{figure*}[ht]
	\centering
	
	\resizebox{0.6\textwidth}{!}{
		\begin{tikzpicture}
		\node[anchor=south west,inner sep=0] at (0,0)
		{\includegraphics[width=1\linewidth]{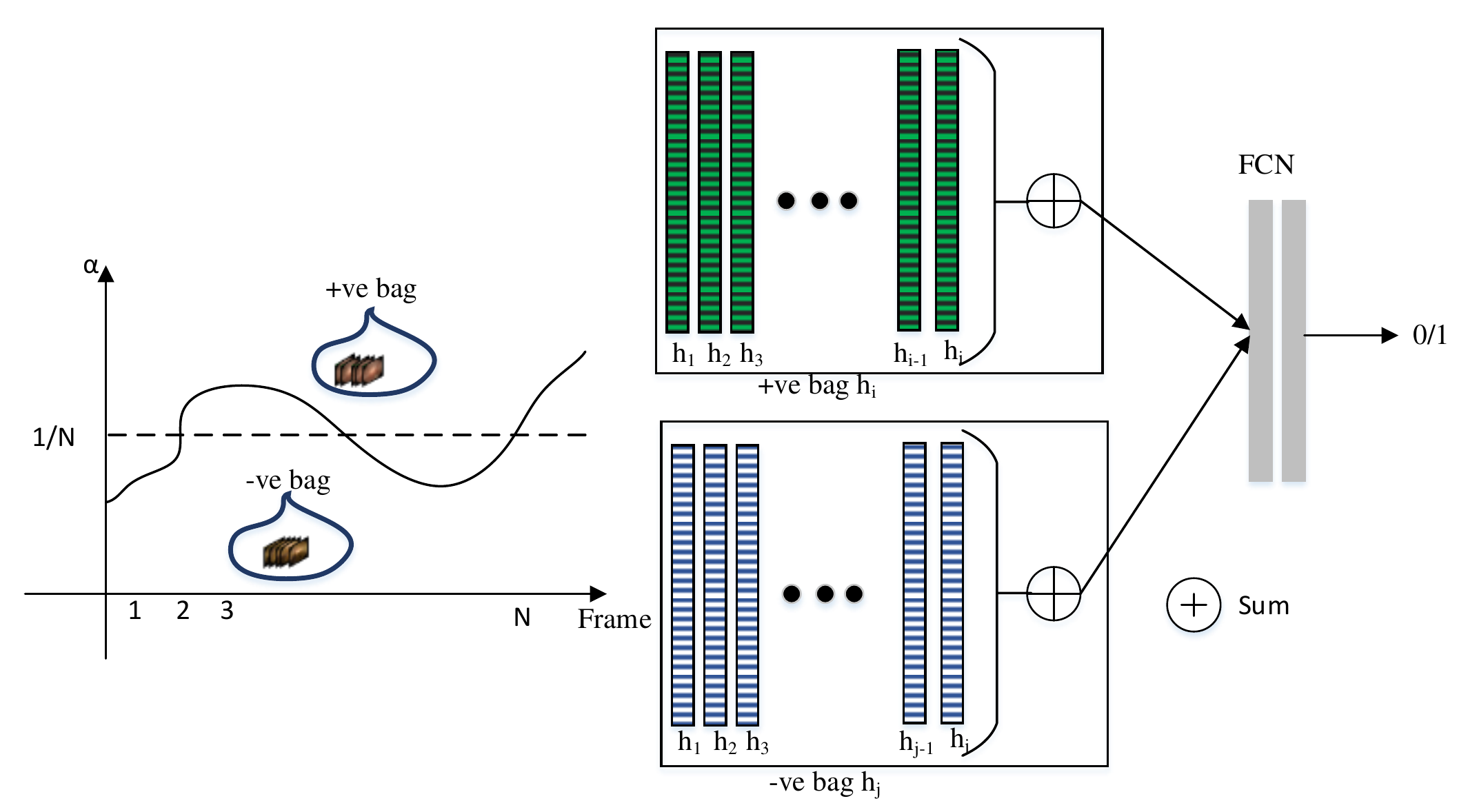}};
		~\node at (13.5,6.0) { $Z^+_{bag}$};
		~\node at (13.5,3.9) { $Z^-_{bag}$};
		\end{tikzpicture}
	}
	\caption{Self-supervision: Frames with high attention values correspond to key frames. Key frames are more likely to contain a given pathology and similar in appearance within a given video. The positive and negative bags are estimated by clustering the frames based on the attention. The aggregation of positive and negative bag embeddings $h_i$ is used for training the self-supervision network together with multi-label video classification network.}
	\label{fig:fig4}
\end{figure*}

Several recent papers have explored the usage of the temporal ordering of images~(\cite{wei2018learning,basha2012photo}).
Using self-supervised training, temporal ordering has been used for representation learning~(\cite{wei2018learning}).
Inspired by temporal ordering, in this work we propose attention ordering for faster convergence and MIL training regularizer.
Our aim is to model how frames with large and small value of attention $\alpha_i \in \boldsymbol{\alpha}$ vary in embedding space $h_i \in \boldsymbol{H}$. Formally, our method of self-supervision works by enforcing the fact that high valued attention $\alpha_i$ aggregated embeddings $h_i$ should be different from low-valued attention  $\alpha_i$ aggregated embeddings.
The self-supervision block is shown in Fig. (\ref{fig:fig4}).
Given frame attention, $\boldsymbol{\alpha}$, the video frames are clustered into positive and negative bags based on the value of $\boldsymbol{\alpha}$ as shown in Eq.(\ref{eq:1added}) and (\ref{eq:1addedneg}).
\begin{equation}
\label{eq:1added}
Z^+_{bag} = \sum_{b=1}^{B^+=|\boldsymbol{\alpha }>\frac{1}{N}|} h^+_b \mid  h^+_b \subseteq \boldsymbol{h}, \alpha_b>\frac{1}{N} 
\end{equation}

\begin{equation}
\label{eq:1addedneg}
Z^-_{bag} = \sum_{b=1}^{B^-=|\boldsymbol{\alpha }\le\frac{1}{N}|} h^-_b \mid  h^-_b \subseteq \boldsymbol{h}, \alpha_b\le\frac{1}{N} 
\end{equation}
where $B^+$ and $B^-$ are the cardinalities of the set $\boldsymbol{\alpha }>\frac{1}{N}$ and $\boldsymbol{\alpha }\le\frac{1}{N}$ respectively.

Finally, positive and negative bag feature embeddings, $Z^+_{bag}, Z^-_{bag}$, are used for training a two-layered neural network. The network is trained with the ground truth value of ``1'' if the bags are the same and ``0'' otherwise. In other words, the proposed self-supervision acts as a regularizer by maximizing the distance between positive and negative bags in embedding space.
\subsection{Loss function}
The inputs of our model consist of a sequence of video frames and their corresponding pathology label (ground truth). Considering we would like to learn both temporal attention and video-level predictions, we formulated the loss function shown in Eq. (\ref{eq:5}).  The purpose of
the training process is to minimize such loss function $\mathcal{L}$, where $M$ is the size of the training set, $g$ are the ground truth labels, and $y$ are the predicted probabilities.
\begin{equation}
\label{eq:5}
\begin{aligned}
\mathcal{L}= \frac{1}{M}\sum_{m=1}^{M} g_m\log(y_m) +  \\ -\frac{\lambda}{M}\sum_{m=1}^{M} g_m^{bag}\log(y_m^{bag})-(1-g_m^{bag})\log(1-y_m^{bag})
\end{aligned}
\end{equation}
The first term part of Eq. (\ref{eq:5}) minimizes video-level prediction loss. Note that unlike~(\cite{sharma2015action,xu2015show}), here the attention is learned implicitly without any constraint in the loss function.  The second part of the equation represents self-supervision loss which is a negative log-likelihood based on the Bernoulli distribution. $g_m^{bag}$ is the ground truth label, which is ``1'' if the bags are the same or ``0'' otherwise. $y_m^{bag}$ is the predicted probabilities of the bag.

\begin{figure*}[ht]
	\centering
	
	\resizebox{0.6\textwidth}{!}{
		\begin{tikzpicture}
		\node[anchor=south west,inner sep=0] at (0,0)
		{\includegraphics[width=1\linewidth]{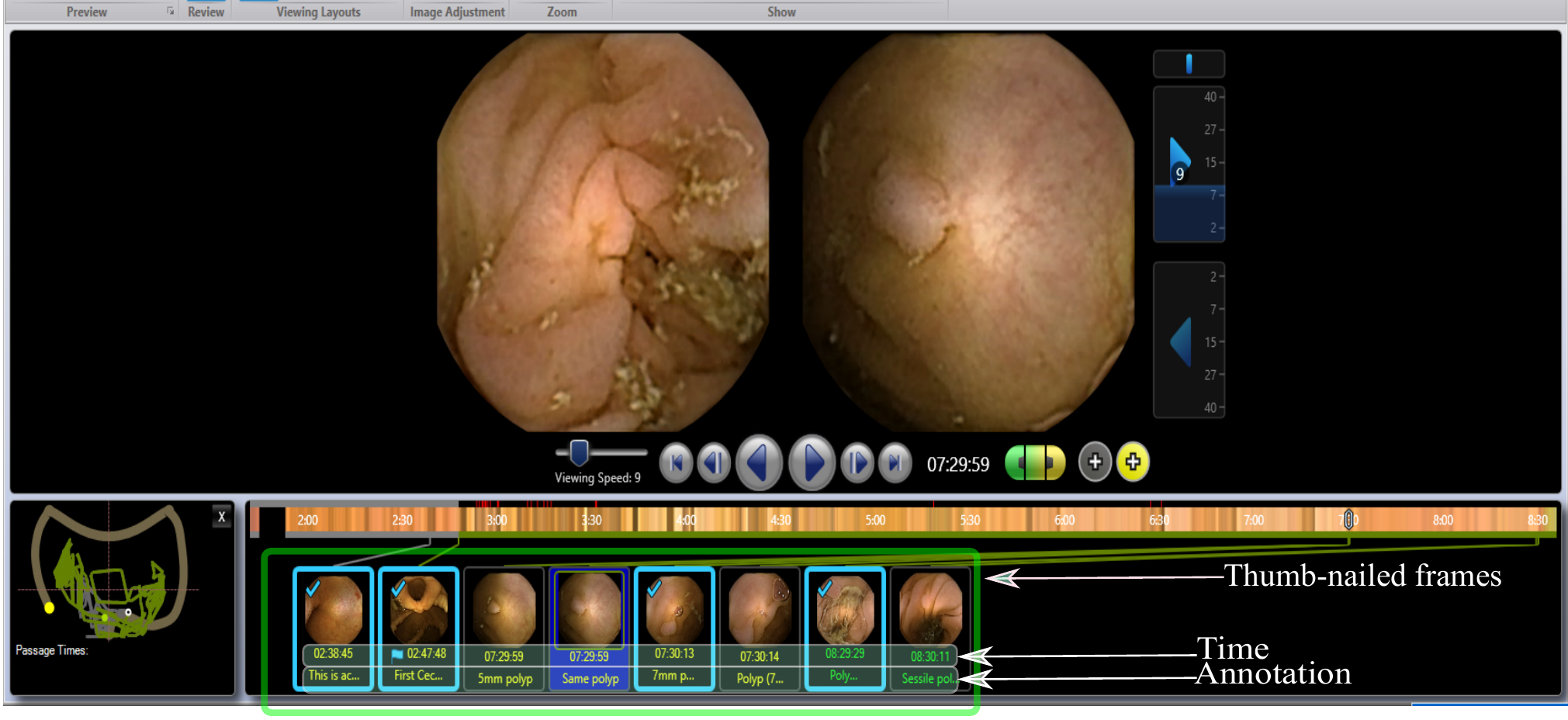}};
		\end{tikzpicture}
	}
	\caption{Annotation: The top right and left images show the rear and front camera view of PillCam COLON II capsule. Gastroenterologist thumb-nail a given instances of a suspected pathology with annotation text as shown above. A video with 50 to 100 frames containing such thumb-nailed frame is extracted as a training and test video.  }
	\label{fig:fig4anno}
\end{figure*}

\section{Experiment}
In this section, we provide an analysis of our proposed PS-DeVCEM model and evaluate our temporal attention method. Then we compare our model with representative state-of-the-art methods~(\cite{ray2005supervised,andrews2003support,zhou2007relation,bunescu2007multiple,ilse2018attention,paul2018w,nguyen2018weakly}) and evaluate them quantitatively for each of the pathologies.\\
\subsection{Dataset}
This work is joint work with a hospital aimed at medical application.
The dataset is collected using PillCam COLON I and II VCE devices from 40 patients. 
PillCam COLON VCE  is 11mm x 31mm in size and it is equipped with two cameras acquiring pictures from both ends of the capsule with an adaptive frame rate of 4-35 frames per second~(\cite{mohammed2018variational}). The dataset consists of 455 short segment videos with a total of  28,304 images. 
We extracted the video segments using the RAPID reader software~(\cite{RAPIDRe77:online}) from the gastroenterologist tagged section of approximately 8-hour video per patient as shown in Fig. (\ref{fig:fig4anno}). 
Each training sample consists of 50 to 100 frames with the middle frame
thumb-nailed by a gastroenterologist. The videos are of 512 $\times$ 512 resolution.
The dataset is labeled by a single doctor and later on verified by a second medical doctor (experienced gastroenterologist). 
The dataset is unbalanced as some pathologies are more frequent than others. 
The dataset is representative of a clinical setting and we kept rare pathologies in the dataset. 
The dataset includes 14 classes showing pathological findings and cleansing quality of the endoscopic procedures, Table \ref{tab:val}.\\

\begin{table*}[]
	\caption{Content of PS-DeVCEM dataset: Note that some of the video segments are labeled for multiple pathologies. Each video is labeled by one gastroenterologist and checked by a second gastroenterologist for quality control. On average, the training and test data has 1.74 and 1.85 labels per video respectively. In the training data number of videos having one, two, three, four, and five labels are 107, 88, 16, 14, and 2 videos respectively. In the test data videos having one, two, three, four, five, six labels are 98, 89, 21, 17, 2, and 1 respectively.} 
	\label{tab:val}
	\centering   
	\resizebox{1\textwidth}{!}{%
		\begin{tabular}{|l|c|c|c|c|c|c|c|c|c|c|c|c|c|c|c|} 
			\hline
			\hline
			\rule[-1ex]{0pt}{3.5ex}  Pathology & Erosions  &  Debris &Diverticulosis&Erythema&Granularity&Haemorrhage&Inflammation&Normal&Oedema&Angioectasia& Polyp&Pseudopolyp&Tumor&Ulceration&\textbf{Total}\\
			\hline
			\rule[-1ex]{0pt}{3.5ex}  \# training videos & $54$ & $72$& $17$& $16$ & $27$& $17$ & $22$ & $45$ & $5$& $1$ & $32$  & $28$ & $8$& $32$& $\boldsymbol{227}$   \\
			\hline
			\rule[-1ex]{0pt}{3.5ex}  \# testing videos  & $64$& $84$& $16$ & $21$& $28$& $20$ & $24$& $41$& $7$  & $1$  & $30$ & $29$ & $3$& $45$ & $\boldsymbol{228}$ \\
			
			\hline
			\hline
	\end{tabular}}
\end{table*} 
\textbf{Dataset splitting:} For proper ablation study and benchmarking, we split the dataset into two groups, train/test. Data splitting can be formulated as a statistical sampling problem. There are various statistical sampling techniques that could be employed to split the data~(\cite{may2010data}). In our case, we used simple random sampling with 50\% of the data for training and 50\% for testing. In such a case, we try to make a train/test set to contain a more or less equal number of pathological findings and artifacts. The train/test set video are sampled randomly from all patients to insure the trained model learned to distinguish the diseases rather than different patients. Table \ref{tab:val} outlines information about the pathologies and the number of videos in the training and testing set.  \\
\textbf{Data augmentation:} We randomly flipped the video segments horizontally or vertically and randomly zoom parts of the video segment to prevent the network from overfitting. We acknowledge that extensive data augmentation techniques (for instance, swapping temporal order, perspective distortion) will likely lead to improved performance. However, since the purpose of this evaluation is to benchmark different methods, we rely on simple data augmentation techniques.\newline
\textbf{Implementation:} Our model is implemented with a Pytorch library with a single NVIDIA TITAN X GPU. The images are resized into fixed dimensions with a spatial size of 224 $\times$ 224 before feeding the encoder networks. The encoder network follows the typical architecture of ResNet50~(\cite{he2016deep}), which has been widely used as the base network in many vision applications. The encoders are shared and initialized with a pre-trained weight trained on the ImageNet dataset. The last fully-connected layer of the network was truncated and the output of average pooling is used for frame representation. We set the sequence length to 30 frames per video segment with a bi-directional LSTM  hidden-state dimension of 1024. Longer videos are sampled uniformly to a constant size of 30 frames. For more details, please refer to Appendix.\\
\textbf{Frame-level inference:} The importance of each frame to the final video level representation is determined by the value of $\alpha$ as shown in Eq. (\ref{eq:4}). Frames with a high value of $\alpha$ indicate where in the video a given pathology is suspected, hence providing frame-level inference. \\
\textbf{Evaluation:} We report our experimental results using the PS-DeVCEM dataset. Following earlier works, \cite{tajbakhsh2015, bernal2017comparative}, evaluation is done using precision, recall, F1-score, and sensitivity metrics. Low recall could lead to miss-diagnosis while low precision could add extra work to the gastroenterologist. Hence, having a high performance with a balanced Type I and II error and preferably lower Type II error would be desirable. In all of our experiments, we kept the base network to ResNet50 and we examined how our PS-DeVCEM approach handles challenging cases of video and frame-level inference. \\
\textbf{Ablation study on temporal attention and self-supervision:} After getting visual feedback from a gastroenterologist on temporal attention accuracy, we performed multiple experiments to improve the accuracy of the attention block. There are various ways to train the temporal attention block based on the input feature used. To analyze the impact of using various input features and the importance of each block in the proposed framework, we carry out extensive ablation studies on the PS-DeVCEM dataset. The results are summarized in Table \ref{tab:ablation}. Table \ref{tab:ablation} shows multiple experiments on temporal attention block positioning and training the network. We evaluate the optimal placement of temporal attention block as follows.  \\
\textit{Learning on frame feature (AttenConv):} The temporal attention block is fed with the extracted feature from each frame $x_i$. The frame representation and temporal attention are given in Eq. (\ref{eq:6}) and (\ref{eq:7}). Each convolution feature is weighted with computed value $\alpha_n$ before feeding into the LSTM network. The final state of the LSTM ($h_N$) is used for training the neural network. This is equivalent to applying temporal attention to extracted features and model the temporal information with LSTM. Therefore, after temporal attention the extracted feature $x_n$ becomes
\begin{equation}
\label{eq:6}
\hat{x}_n= x_n\alpha_n
\end{equation}
where
\begin{equation}
\label{eq:7}
\alpha_n= \frac{\exp\{\boldsymbol{w}^T \text{tanh}(\boldsymbol{Vx}_k^T)\}}{\sum_{n=1}^{N} \exp\{\boldsymbol{w}^T \text{tanh}(\boldsymbol{Vx}_k^T)\} }
\end{equation}\\
\textit{Learning on frame feature for LSTM attention (AttenConvLSTM):} Another alternative to AttenConv configuration is to train the attention module on the input frame feature while weighting the hidden state of LSTM block with the computed attention weights. Hence, for this configuration, the MIL pooling layer is given by Eq. (\ref{eq:3}) and the temporal attention weights are calculated with Eq. (\ref{eq:7}). In other words, each frame contribution to the final video-level representation is determined with extracted feature $x_i$ without any temporal information. \\ 
\textit{Learning attention on hidden states of LSTM (AttenLSTM):} This is a typical approach of computing attention with LSTM blocks for human action recognition~(\cite{chan2018human,sharma2015action,xu2015show}) tasks. In this case, the temporal attention is computed using the hidden state representation of LSTM $h_i$ and extracted feature $x_i$. The MIL pooling layer is given by Eq. (\ref{eq:3}) with temporal attention weights as shown in Eq. (\ref{eq:4}). The difference between this configuration and the proposed method is, in PS-DeVCEM we use the residual block shown in Fig. (\ref{fig:fig2}) and self-supervision Fig. (\ref{fig:fig4}).\\
\textit{Guided AttenLSTM (GuidedLSTM):} In this configuration, we extended our earlier experiment, AttenLSTM by introducing the self-supervision network that is introduced in Section \ref{sec:selfsuper}. The self-supervision block is trained to minimize the distance between high and low attention weighted frame feature representation. The main purpose of this experiment is to examine the efficacy of self-supervision on the overall accuracy of the method and temporal attention. 
\begin{table*}[t]
	\caption{Ablation study result: The values are averaged for all pathologies. The table shows different ways of computing attention weights and its impact on the overall performance of video classifiers. With the proposed residual LSTM block and self-supervision, the final video representation gives better performance.  } 
	\label{tab:ablation}
	\centering      
	\begin{tabular}{lcccccc} 
		\hline
		\hline
		\rule[-1ex]{0pt}{3.5ex}  Method  & Precision& Recall& F1-score& Specificity \\
		\hline
		\rule[-1ex]{0pt}{3.5ex} AttenConv   &  0.229  &  0.290  &  0.246      &  0.872 \\  
		\hline
		\rule[-1ex]{0pt}{3.5ex} AttenConvLSTM   &  0.450   & 0.461  &  0.443   &  0.939    \\  
		\hline
		\rule[-1ex]{0pt}{3.5ex} AttenLSTM   &  0.529  & 0.478  &  0.487    &  \textbf{0.954 }     \\  
		\hline
		\rule[-1ex]{0pt}{3.5ex} GuidedLSTM  &  0.487  & 0.482  &  0.458    &  0.946       \\  
		\hline
		\rule[-1ex]{0pt}{3.5ex} PS-DeVCEM(proposed)   &  \textbf{0.616}  & \textbf{0.546}  &  \textbf{0.551}     &  0.951     \\  
		\hline
		\hline        
	\end{tabular}
\end{table*}
\begin{figure*}[!htbp]
	\begin{subfigure}{1\textwidth}
		\centering
		\includegraphics[width=1\linewidth]{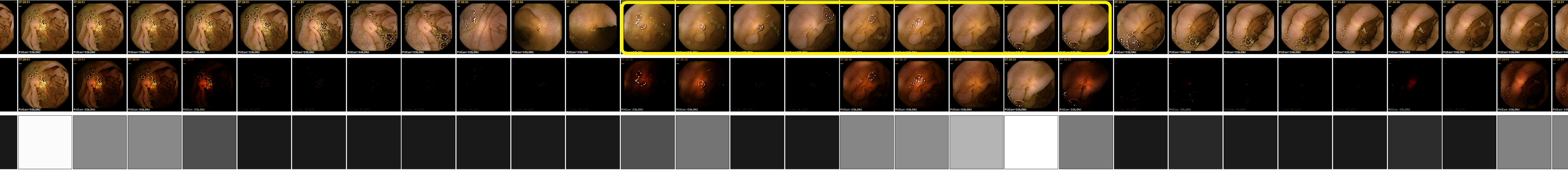}
		\caption{AttenConv attention. Predicted:``Polyp''}
		\label{fig:convconvab}
	\end{subfigure}\\
	\begin{subfigure}{1\textwidth}
		\centering
		\includegraphics[width=1\linewidth]{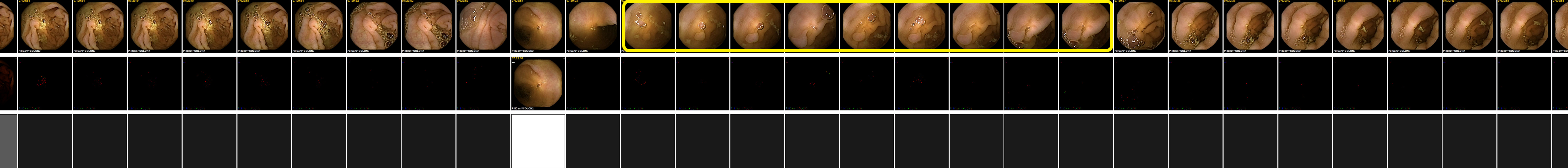}
		\caption{AttenConvLSTM attention. Predicted: ``Polyp''}
		\label{fig:ConvLSTM}
	\end{subfigure}\\
	\begin{subfigure}{1\textwidth}
		\centering
		\includegraphics[width=1\linewidth]{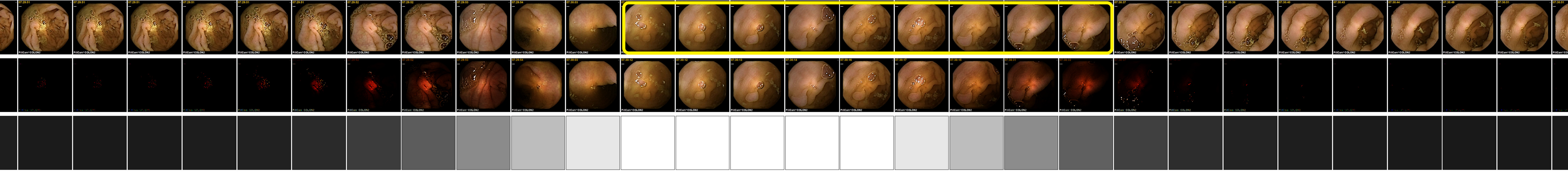}
		\caption{AttenLSTM attention. Predicted: ``Polyp''}
		\label{fig:AttenLSTM}
	\end{subfigure}
	\begin{subfigure}{1\textwidth}
		\centering
		\includegraphics[width=1\linewidth]{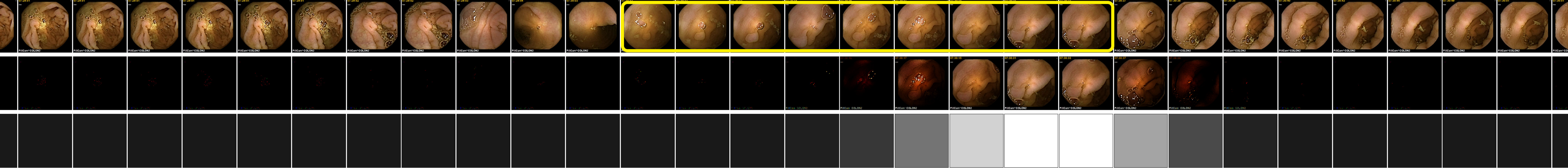}
		\caption{GiudedLSTM attention. Predicted: ``Polyp''}
		\label{fig:GiudedLSTM}
	\end{subfigure}
	\begin{subfigure}{1\textwidth}
		\centering
		\includegraphics[width=1\linewidth]{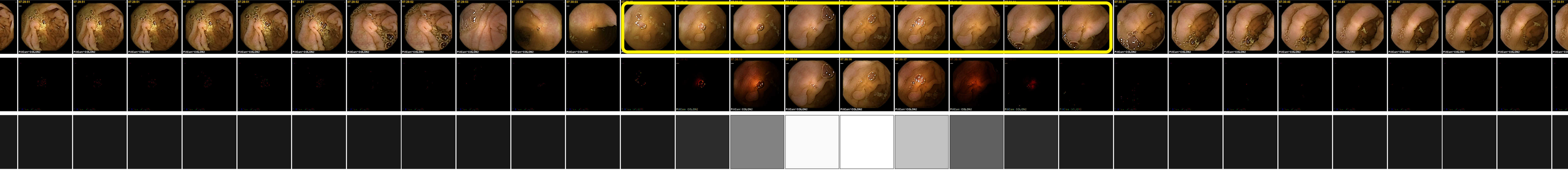}
		\caption{PS-DeVCEM(proposed) attention. Predicted: ``Polyp''}
		\label{fig:PS-DeVCEM}
	\end{subfigure}
	\caption{Example of attention weights for different configurations. The top frames show the sequence of video frames with the corresponding attention in the middle frame. Yellow bounding boxes on the top frames show frame level annotation. The last row shows gray scale coded image with black the irrelevant frames and white the relevant frames. Attention frames vary from Blank (Black) to the original frame corresponding to low and high value of attention weights, respectively. The attention images are encoded as $(I^{ 0.0001 +\frac{1}{attention}})$. The ground truth label for the video is ``Polyps" shown in yellow box, with expected attention to the middle and left side of the video. It can be seen that our proposed method gives smooth attention as well as better localization of suspected frames.(Best viewed in color)}
	\label{fig:comp_atten}
\end{figure*}\\
\textbf{Ablation study results:}
Table \ref{tab:ablation} lists the results of the four variants of our framework discussed along with the temporal attention weights shown in Fig. (\ref{fig:comp_atten}). From the experiments, we note the following points for improved trainable MIL pooling layer. First, learning attention weights from LSTM hidden state gives a better result compared to convolution features. Although VCE videos are taken with an adaptive frame rate of 4-35 frames per second, temporal information helps in improving the overall performance of the networks. Moreover, as shown in Fig. (\ref{fig:comp_atten}), attention weights are not smooth for visually similar neighboring frames.  Secondly, even with temporal information learning attention from convolutional features $x_i$ alone performs worse than learning attention weights with LSTM hidden states. Thirdly, our method with residual features and self-supervision gives a better result with 8.7\% and 6.8\%  improvement in average precision and average recall as compared to second-best metrics for each metric respectively. It is important to note that the above experiments are done under the same experimental setup with different configuration outlined above.

\begin{figure*}[!htbp]
	\begin{subfigure}{0.33\textwidth}
		\centering
		\includegraphics[width=1\linewidth]{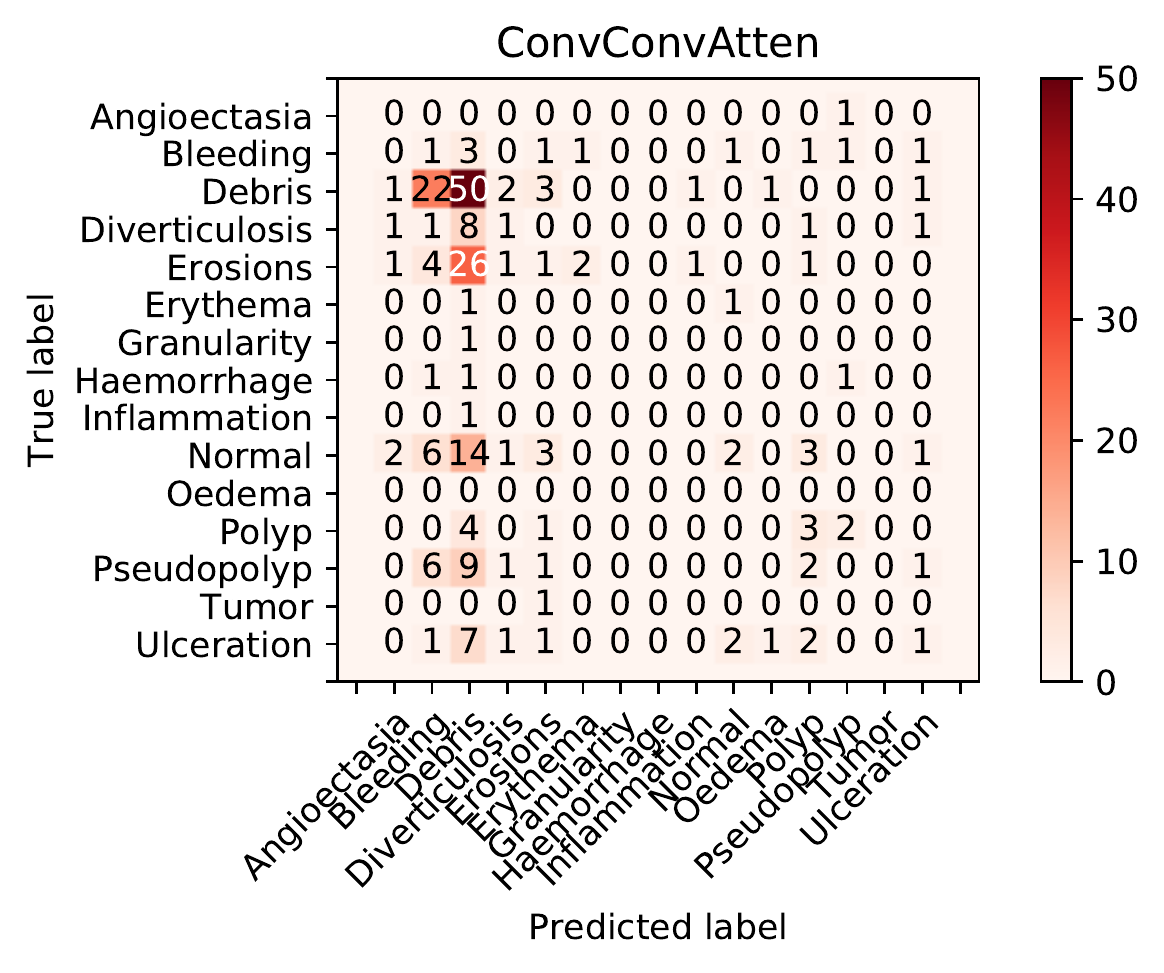}
		\caption{AttenConv}
		\label{fig:convconv_conf}
	\end{subfigure}
	\begin{subfigure}{0.33\textwidth}
		\centering
		\includegraphics[width=1\linewidth]{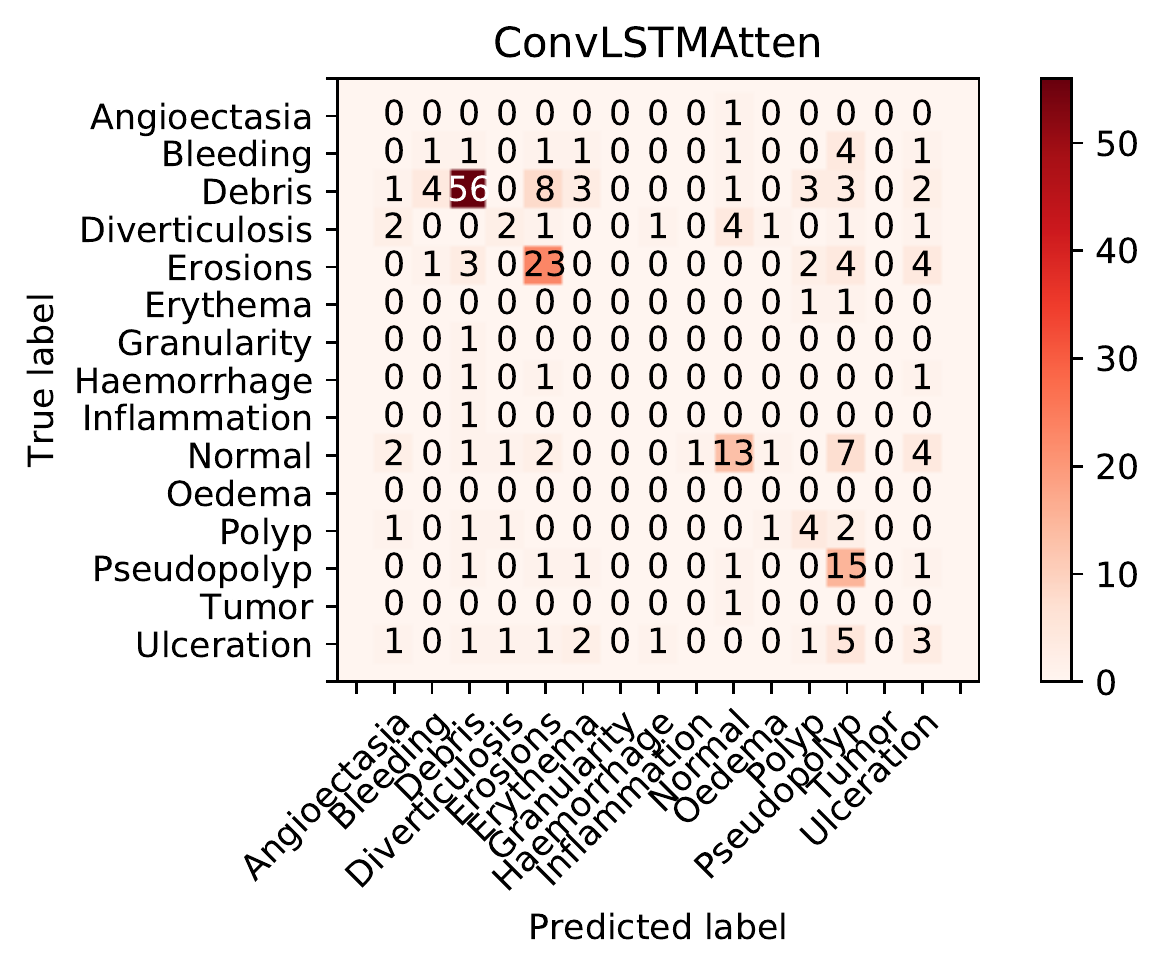}
		\caption{AttenConvLSTM}
		\label{fig:ConvLSTM_conf}
	\end{subfigure}
	\begin{subfigure}{0.33\textwidth}
		\centering
		\includegraphics[width=1\linewidth]{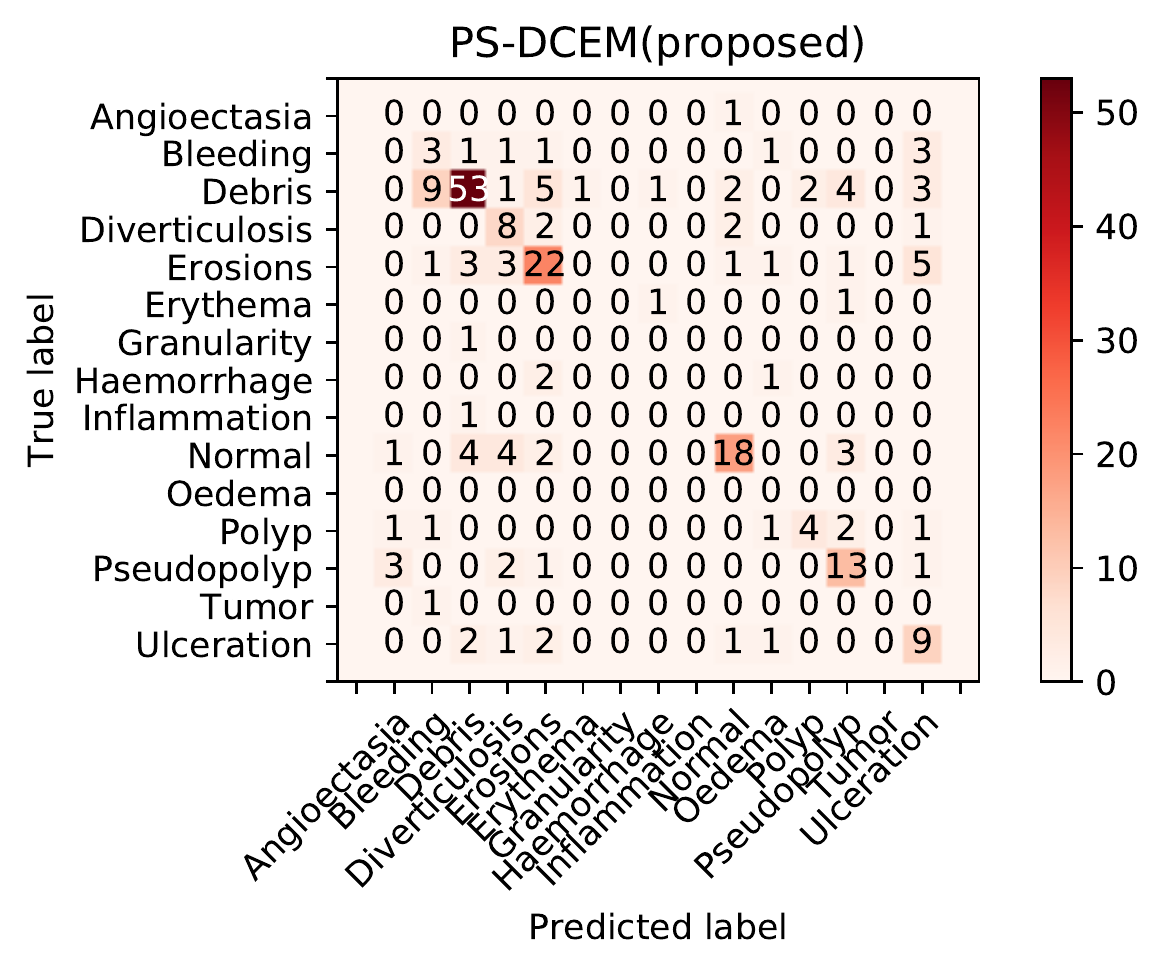}
		\caption{PS-DeVCEM(proposed)}
		\label{fig:AttenLSTM_conf}
	\end{subfigure}
	\caption{Learning attention: In Fig. (\ref{fig:convconv_conf}), the video feature is computed by using the weighted sum of features $x_i$ using the attention weights computed based on the convolutional feature while in Fig. (\ref{fig:ConvLSTM_conf}) and (\ref{fig:AttenLSTM_conf}) is computed using the output of the residual LSTM block. In Fig. (\ref{fig:convconv_conf}), we can see that normal frames are confused for bleeding, debris and diverticulosis as compared to the proposed method. Furthermore, with our proposed self-supervision very similar classes such as ulcerations and erosion are better separated. In the above figure, Fig. (\ref{fig:convconv_conf}) represents independent sample based method while Fig. (\ref{fig:ConvLSTM_conf}) and (\ref{fig:AttenLSTM_conf}) represent temporal based MIL formulations as discussed in related work section.}
	\label{fig:comp_atten_conf}
\end{figure*}

\begin{figure*}[htbp]
	\begin{subfigure}{1\textwidth}
		\centering
		\includegraphics[width=1\linewidth]{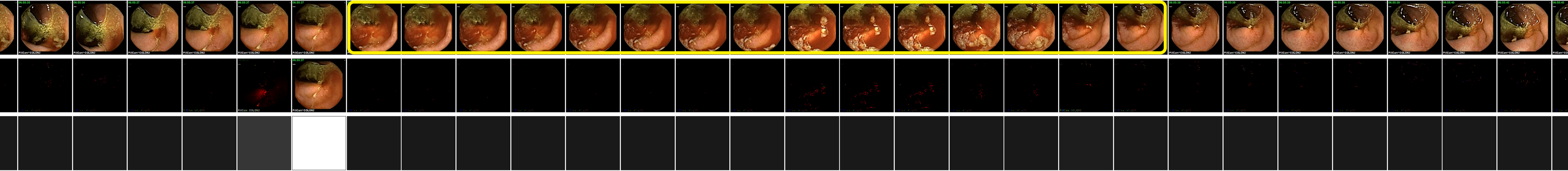}
		\caption{Attention weights~(\cite{ilse2018attention}), Predicted: ``Bleeding''}
		\label{fig:convconv}
	\end{subfigure}\\
	\begin{subfigure}{1\textwidth}
		\centering
		\includegraphics[width=1\linewidth]{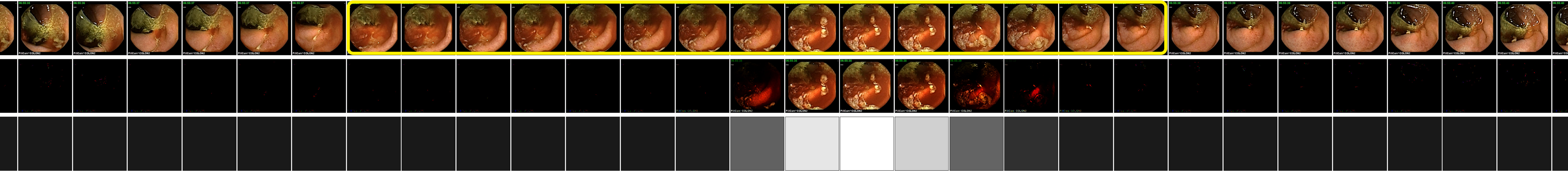}
		\caption{PS-DeVCEM(proposed) attention.  Predicted: ``Bleeding''}
		\label{fig:ConvLSTM}
	\end{subfigure}\\
	\caption{Comparison of temporal attention weights. The ground truth label for the video is ``Bleeding", in most of the frames. In~(\cite{ilse2018attention}), it is assumed that each instance is permutation-invariant and the attention modules are not able to localize the keyframes. Our approach considers neighboring instances to be similar and therefore gives a smooth and better localization of the keyframes.}
	\label{fig:sota_atten}
\end{figure*}

Temporal attention weights are shown in Fig. (\ref{fig:comp_atten}) examine contributions from each frame to the final video contribution.
The attention weights give important insight into the location of the pathology in the video.
Note that, since the attention weights are normalized, the value corresponds to the probability of each frame to have a given pathology.
Although GuidedLSTM performs slightly less than AttenLSTM, the attention weights are narrower as they are more discriminative between positive and negative classes. On the other hand self-supervision with residual blocks and AttenLSTM attends the correct frames as compared to other methods. (i.e. few frames on the left and middle of the video shows polyp) \\
\textbf{Comparison with state-of-the-art:} In this subsection, we present our experimental results with classical and recent MIL works using the PS-DeVCEM dataset. In order to compare with~(\cite{ray2005supervised,bunescu2007multiple,zhou2007relation}), we extracted image features with ResNet50 architecture for similar representation with other methods. It is important to note that these methods are proposed for binary cases. Therefore, we apply the algorithms for each class of pathologies and solve for binary MIL formulation. We also compare with deep learning-based approaches~(\cite{ilse2018attention,paul2018w,nguyen2018weakly}). The feature extraction part of~(\cite{ilse2018attention,paul2018w,nguyen2018weakly}) is replaced with ResNet50 for uniform feature representation. The loss function in~(\cite{ilse2018attention}) is modified to multi-label classification problem. Table \ref{tab:sota} shows that the proposed method does achieve a state-of-the-art result on the PS-DeVCEM dataset.  
All deep learning-based methods~(\cite{ilse2018attention,paul2018w,nguyen2018weakly}) are trained end-to-end with ResNet50 as a backbone network for feature extraction. 
It is important to note that both \cite{nguyen2018weakly} and \cite{paul2018w} are weakly supervised works that are proposed for activity recognition in a video. 
For \cite{nguyen2018weakly} we used a segment size of one and in \cite{paul2018w} we only used the RGB stream.
As in Table \ref{tab:sota}, it is clear that the proposed PS-DeVCEM improves F1-score and precision when using residual LSTM blocks and self-supervision. However, in special cases where pathology exists throughout the video Fig. (\ref{fig:sota_atten}), our proposed method underestimates the frame attention weight (i.e. the video frames are visually similar but the attention values tend to be different). This is due to the dataset imbalance in the training examples for each pathology.
\begin{table*}[t]
	\caption{Comparison with other MIL algorithms: The values are averaged for all pathologies. Note that STPN~(\cite{nguyen2018weakly}) , W-TALC~(\cite{paul2018w}) and Attention based deep MIL~(\cite{ilse2018attention}) are deep neural network based methods while SIL~(\cite{ray2005supervised}) and MissSVM~(\cite{zhou2007relation}) are based on SVM classifier. } 
	\label{tab:sota}
	\centering       
	\begin{tabular}{lcccccc} 
		\hline
		\hline
		\rule[-1ex]{0pt}{3.5ex}  Method    & Precision& Recall& F1-score& Specificity \\
		\hline
		\rule[-1ex]{0pt}{3.5ex} SIL~(\cite{ray2005supervised})   &       0.235 &       0.046  &        0.066  &           \textbf{0.997}    \\  
		\hline
		\rule[-1ex]{0pt}{3.5ex} MissSVM~(\cite{zhou2007relation}) &  0.130  & 0.162  &  0.123    &  0.912     \\  
		\hline
		\rule[-1ex]{0pt}{3.5ex} Attention based deep MIL~(\cite{ilse2018attention}) &   \textbf{0.616}  & 0.471  &  0.513     &  0.955          \\  
		\hline
		\rule[-1ex]{0pt}{3.5ex} STPN~(\cite{nguyen2018weakly})   &       0.592 &       0.517  &        0.536     &       0.916   \\  
		\hline
		\rule[-1ex]{0pt}{3.5ex} W-TALC~(\cite{paul2018w})  &  0.274  & \textbf{0.891}  &  0.416   &  0.666    \\  
		\hline
		\rule[-1ex]{0pt}{3.5ex} PS-DeVCEM(w/o self-supervision)  & 0.606 &0.54  &    0.54    &  0.951     \\  
		\hline
		\rule[-1ex]{0pt}{3.5ex} PS-DeVCEM(proposed)&  \textbf{0.616}  & 0.546  &  \textbf{0.551}      &  0.951     \\  
		\hline
		\hline
		
	\end{tabular}
\end{table*}

\begin{figure*}[!htbp]
	\begin{subfigure}{0.5\textwidth}
		\centering
		\includegraphics[width=1\linewidth]{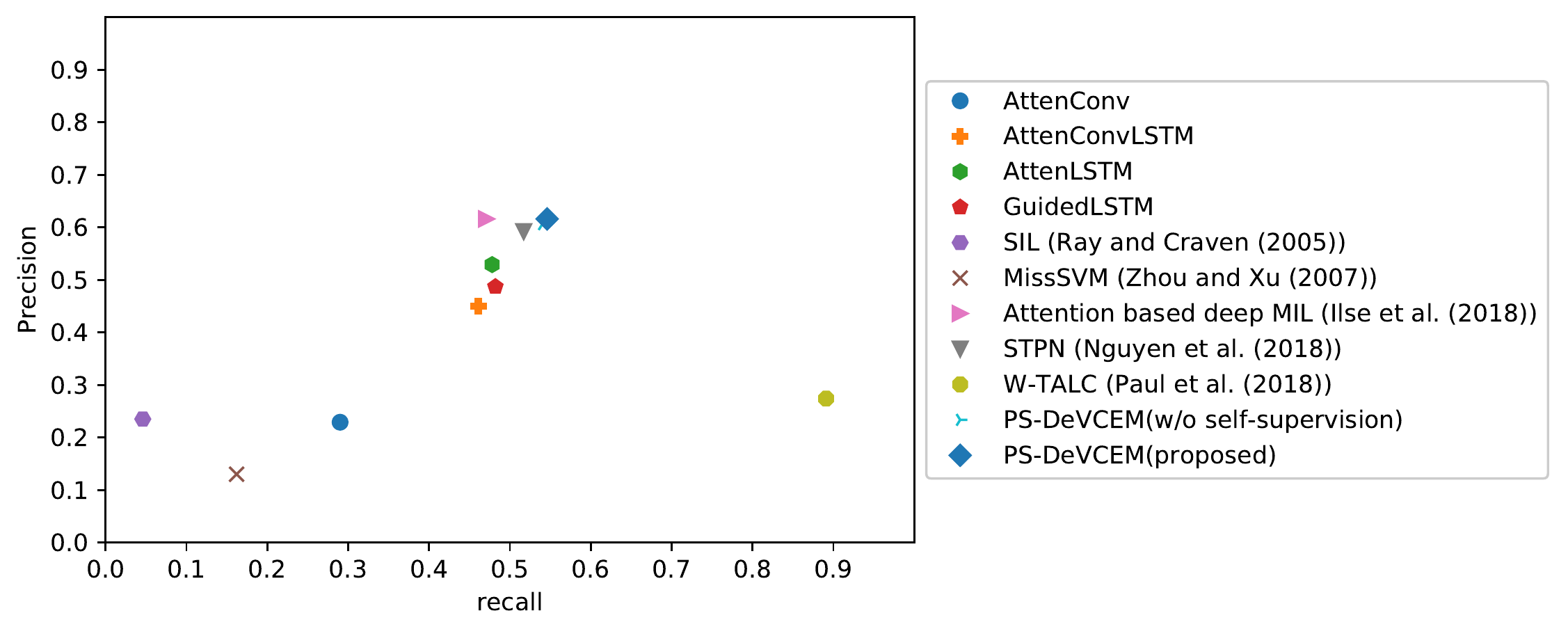}
		\caption{Precision-recall scatter plot for the proposed model and some state of art methods on capsule endoscopy dataset. }
		\label{fig:prec_recal}
	\end{subfigure}
	\begin{subfigure}{0.5\textwidth}
		\centering
		\includegraphics[width=1\linewidth]{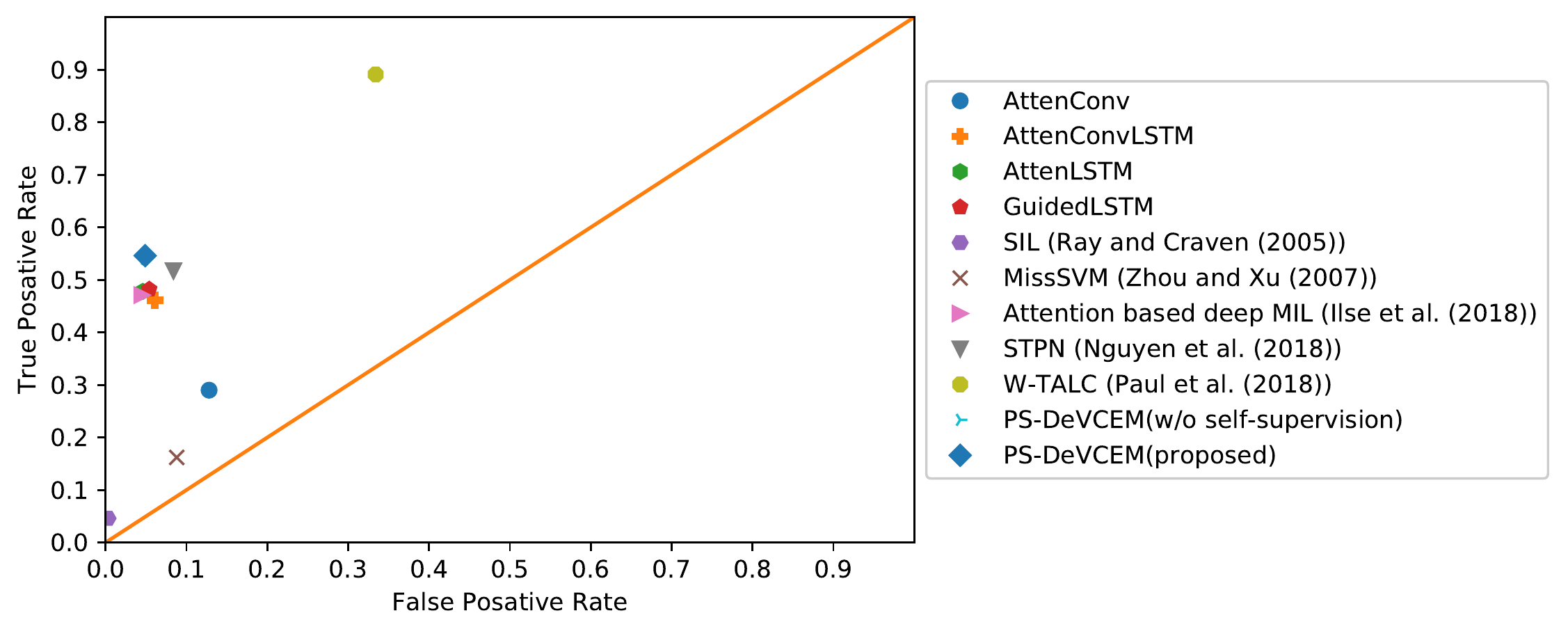}
		\caption{ROC scatter plot.}
		\label{fig:roc}
	\end{subfigure}
	\caption{Performance evaluation on multi-label video classification. The plot shows comparison of state of the art methods with different ways of computing the attention weights. From precision-recall plot, we can see that W-TALC~(\cite{paul2018w}) gives a high recall value and small precision score compared to the proposed method. However, low precision score results in additional work for the gastroenterologist as false positive sample videos need to be reviewed.}
	\label{fig:prec_roc}
\end{figure*}
\textbf{Self-supervision:} In order to understand how self-supervision loss affect the detection performance, we have included further experiment in Table \ref{tab:sota}.
As shown in Table \ref{tab:sota}, self-supervision improves the overall performance of the network by maximizing the distance between positive and negative feature embeddings.
The self-supervision allows frames with similar feature to have similar attention weights and regularizes the attention weights to be consistent.\\
\textbf{Discussion:} By using self-supervision and residual LSTM blocks, we effectively optimized the performance of the proposed approach. By first classifying group of frames into positive and negative classes and further classifying the frames as a whole into separate categories progressively, the self-supervision mechanism improves feature discrimination between similar classes. Compared to metric learning techniques such as~\cite{paul2018w} as shown in Table \ref{tab:sota}, the proposed method relies on a weak supervision to improve positive and negative class feature representations.  Alternatively approaches to the self-supervision mechanism including metric learning and siemese network \cite{bromley1994signature} variants can be used with the caveat that each video could contain multiple pathologies and some pathologies are more-likely to occur together than others. Furthermore, since such approach give a high self-supervision, dataset imbalance and representation learning need to be taken into consideration. From the confusion matrix plot in Fig. (\ref{fig:comp_atten_conf}), we can observe that similar classes like debris and erosion, erosion and ulceration are challenging to visually separate. With the proposed weak self-supervision, we are able to improve discriminative feature representation without directly addressing class imbalance problem. However, we note that the frame-level inference could be influenced depending on the following points. Firstly, the dataset is collected with a central part of the video tagged for pathologies. This could influence the learning process in practical settings since it can bias the learning algorithm to memorize the location of the tagged pathology. Secondly, the residual LSTM blocks aggregate information temporally which could miss-align the attention to an incorrect segment of the video. Higher attention weight could be given to frame location where the highest temporal information available. One approach to address the above issues is to collect additional datasets and longer sequences. However, despite being trained in a purely weakly-supervised manner, our approach gives the state-of-the-art result for pathology detection.

As shown in Fig. (\ref{fig:prec_roc}), temporal information aggregation using attention units improves the over all performance of any of methods. However, the method and input to the attention units affect the performance video classification task as well as frame localization. Experimental results as shown in the ablation study indicate that residual LSTM blocks as input to a two layered neural network attention units give a better performance compared to alternative approaches.

\section{Conclusion}
In this work, we proposed PS-DeVCEM: a pathology-sensitive end-to-end deep model based on weakly labeled capsule endoscopy data. We introduced a self-supervision method and residual LSTM blocks for video and frame-level prediction, further improving the interpretability of the proposed framework. Furthermore, we developed the first VCE dataset with video labels aiming at MIL formulation with a total of 455 short-segment videos.  Moreover, experimental results on the PS-DeVCEM database show that the proposed method achieves the best performance on precision and F1-score metrics. Finally, we believe that the PS-DeVCEM dataset and the proposed approach will inspire similar works as the dataset and code will be available with this publication. 

As future work, we plan to improve the video frame localization through domain knowledge of the pathologies. Moreover, some pathologies such as inflammations have longer temporal dependencies and handling longer temporal dependencies can further improve the performance. Furthermore, we are planning to diversify our dataset and collect more videos to improve the frame-level localization of pathologies.

\bibliographystyle{model2-names}
\bibliography{refs}
\newpage

\appendix
\section{Training Details:}

\begin{table}[!htbp] 
	\caption{Training configuration for all deep learning based methods when applicable} 
	\label{tab:appconf}
	\centering 
		\resizebox{0.4\textwidth}{!}{      
	\begin{tabular}{ll} 
		\hline
		\hline
		\rule[-1ex]{0pt}{3.5ex}  Configuration & Description  \\
		\hline
		\rule[-1ex]{0pt}{3.5ex}  Input video (Bag) size & $30 \times 224 \times 224 \times 3$ \\
		\hline
		\rule[-1ex]{0pt}{3.5ex}  Batch size & $1$   \\
		\hline
		\rule[-1ex]{0pt}{3.5ex}  ResNet50 output feature size & $2048$   \\
		\hline
		\rule[-1ex]{0pt}{3.5ex}  Number of Hidden Bidirectional LSTM & $2$ \\
		\hline
		\rule[-1ex]{0pt}{3.5ex}  Hidden Bidirectional LSTM size & $512$ \\
		\hline
		\rule[-1ex]{0pt}{3.5ex}  Detection threshold  & 0.5 \\
		\hline
		\hline
	\end{tabular}}
\end{table}
\begin{table}[!htbp] 
	\caption{Optimization procedure details} 
	\label{tab:appopt}  
			\resizebox{0.9\textwidth}{!}{  
	\begin{tabular}{lllllll} 
		\hline
		\hline
		\rule[-1ex]{0pt}{3.5ex}  Experiment & Optimizer & Coef. RA $\beta$ & Leaning rate & Weight decay & Epochs & Stopping criteria  \\
		\hline
		\rule[-1ex]{0pt}{3.5ex}  All & Adam & $\beta= (0.9,0.999)$  &  (0.0001)Cyclic learning rate & 0.0001 & 500 & lowest validation error  \\
		\hline 
		\hline   
	\end{tabular}}
\end{table}
\begin{table}[!htbp] 
		\caption{MissSVM~(\cite{zhou2007relation}) configuration} 
		\label{tab:appopt}	
		\centering 
				\resizebox{0.4\textwidth}{!}{
			\begin{tabular}{llll} 
			\hline
			\hline
			\rule[-1ex]{0pt}{3.5ex}  Method & Kernel & Regularization& Max-iteration  \\
			\hline
			\rule[-1ex]{0pt}{3.5ex}  MissSVM & RBF & $1$  &  $100$ \\
			\hline
			\hline
		\end{tabular}}
\end{table}	
\begin{table}[!htbp] 

		\caption{SIL ~(\cite{ray2005supervised}) configuration} 
		\label{tab:appopt}
		\centering 
		\resizebox{0.4\textwidth}{!}{
		\begin{tabular}{llll} 
			\hline
			\hline
			\rule[-1ex]{0pt}{3.5ex}  Method & Kernel & Regularization& Scale  \\
			\hline
			\rule[-1ex]{0pt}{3.5ex}  SIL~(\cite{ray2005supervised})  & Linear & $10$ &False \\
			\hline
			\hline
	\end{tabular}}
\end{table}	

\lstset{
	tabsize = 4, 
	showstringspaces = false, 
	numbers = left, 
	commentstyle = \color{green}, 
	keywordstyle = \color{blue}, 
	stringstyle = \color{red}, 
	rulecolor = \color{black}, 
	basicstyle = \small \ttfamily , 
	breaklines = true, 
	numberstyle = \tiny,
}

\begin{lstlisting}[language = Python , frame = trBL , firstnumber = last ,linewidth=13.4cm,caption={PS-DeVCEM: Network configuration}] 
ResNet50
AvgPool2d(kernel_size=7, stride=1, padding=0)
(lstm): LSTM(2048, 512, num_layers=2, bidirectional=True)
attention: 
(0): Linear(in_features=1024, out_features=256, bias=True)
(1): Tanh()
(2): Linear(in_features=256, out_features=1, bias=True)
classifier: 
(0): Linear(in_features=1024, out_features=15, bias=True)
(1): Sigmoid()
bagclassifier: 
(0): Linear(in_features=1024, out_features=1, bias=True)
(1): Sigmoid()
\end{lstlisting}

\begin{figure*}[!htbp]
	\begin{subfigure}{0.33\textwidth}
		\centering
		\includegraphics[width=1\linewidth]{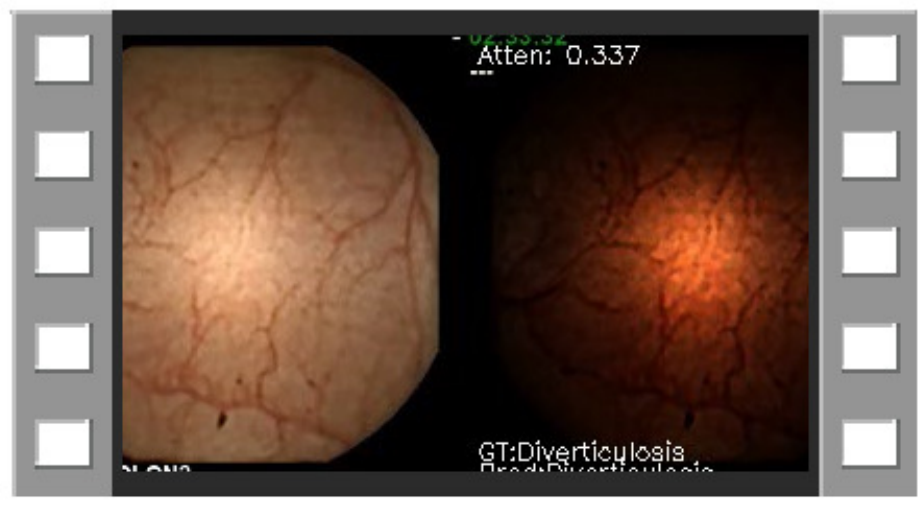}
		\caption{sup1.avi (Attached video): }
		\label{fig:sup1}
	\end{subfigure}
	\begin{subfigure}{0.33\textwidth}
		\centering
		\includegraphics[width=1\linewidth]{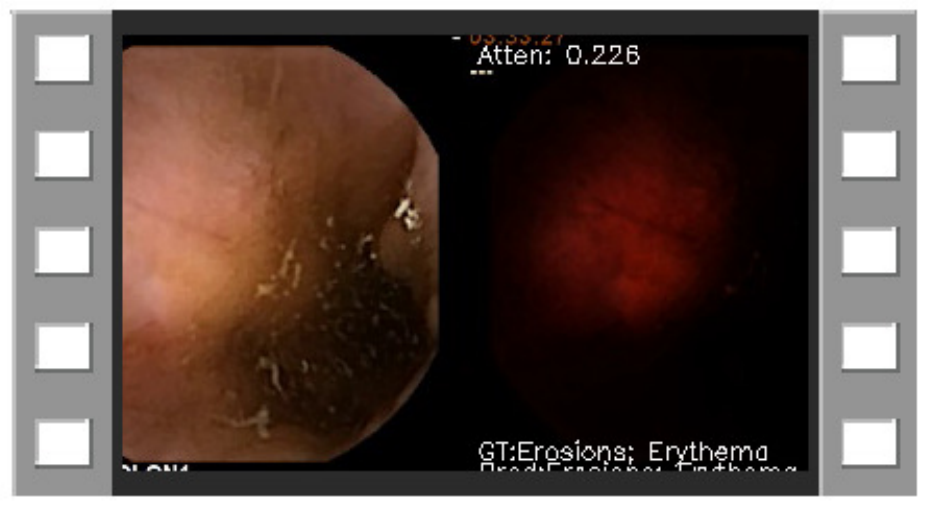}
		\caption{sup2.avi (Attached video):  }
		\label{fig:sup2}
	\end{subfigure}
	\begin{subfigure}{0.33\textwidth}
		\centering
		\includegraphics[width=1\linewidth]{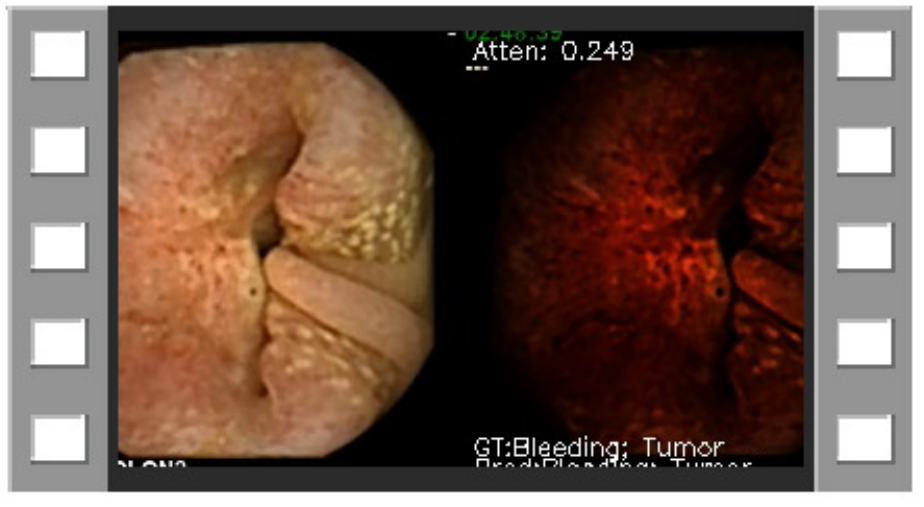}
		\caption{sup3.avi (Attached video): }
		\label{fig:sup3}
	\end{subfigure}
	\caption{We have attached sample videos with  original video segment on the left side. The right side video shows the attention output. Dark frames represent low attention and original frame represents hight attention value. The attention values are shown at the top. Fig. (\ref{fig:sup1}) shows a video with multiple Diverticulosis on different frames. In this case, our method is able to localize the disease from the video labels. Fig. (\ref{fig:sup2}) depicts Erosion and Erythema. Although the prediction for the video is correct, the network attention did not span all the frames with the diseases. Fig. (\ref{fig:sup3}), the segment shows Tumor and Bleeding. As the video shows the Bleeding is not detected with our method. However, some part of the segment containing tumor were localized.}
	\label{fig:videoresult}
\end{figure*}

\end{document}